\documentclass[11pt]{article}
\usepackage[final]{acl}
\usepackage{times}
\usepackage{latexsym}
\usepackage{booktabs}
\usepackage{amsmath}
\usepackage[T1]{fontenc}
\usepackage[utf8]{inputenc}
\usepackage{microtype}
\hypersetup{pdftitle={Thakaa at StanceEval-2026: Inference-Time Target Grounding for Arabic Stance Detection},pdfauthor={Meshal Alamr, Hassan Alqaeri, Abdullah Aldahlawi}}

\newcommand{\favgtwo}{\ensuremath{F_{\mathrm{avg2}}}}
\newcommand{\favgthree}{\ensuremath{F_{\mathrm{avg3}}}}

\title{Thakaa at StanceEval-2026: Inference-Time Target Grounding for Arabic Stance Detection}

\author{Meshal Alamr \quad Hassan Alqaeri \quad Abdullah Aldahlawi \\
  Thakaa, Advanced Company for Information Technology \\
  Riyadh, Saudi Arabia \\
  \texttt{\{m.alamr,h.alqaeril,aldahlawi\}@thakaait.net}}

\begin{document}
\maketitle

\begin{abstract}
StanceEval-2026 tests Arabic stance classification under target shift, with evaluation targets absent from the labelled task data. We test two ways to supply target information at inference time. For Track 1, we retrieve eight labelled Women Driving demonstrations from the external ArabicStanceX corpus and score the three label tokens with Gemma-4-31B-IT. For the unseen-target track, we prepend cached target context prompted to be factual and non-evaluative, then fuse label predictions and label-token scores from multiple LLMs, with final assignments constrained by published class totals. The systems obtain 85.22 and 91.78 \favgtwo{} on Tracks 1 and 2, ranking sixth and fourth in the final standings. On held-out ArabicStanceX tweets, direct prompting scores 79.45 \favgtwo{}; eight random Women Driving demonstrations raise it to 84.78, and similarity retrieval reaches 89.04. Across three target-held-out Mawqif evaluations, a fixed target card raises Gemma-4-31B-IT from 86.54 to 86.93, whereas QLoRA lowers Gemma-4-12B-IT from 84.43 to 83.77. In these separate public evaluations, same-target demonstrations and retrieval improve performance, factual target context provides small, target-dependent gains, and the tested source-only adapter does not improve.
\end{abstract}

\section{Introduction}

Stance detection relates a text to a specified target. Sentiment alone is
insufficient when a tweet criticises an implementation but supports its goal,
or praises an actor without endorsing the target. Arabic social media adds
dialectal variation, sarcasm, and implicit reference
\citep{alturayeif2022mawqif,albalawi2026mawqifv2}.

StanceEval-2026 \citep{albalawi2026stanceeval} evaluates this relation under a
target shift. Track 1 is nominally the seen-target setting, yet it tests Women
Driving after training on
Women Empowerment and two other topics. Its unseen-target track tests Electric
Cars, a transport-technology and industrial-policy topic, and the Trimester
System, a debate about Saudi school-calendar reform. The
Mawqif-XT dataset was designed around these two transfer settings
\citep{albalawi2026mawqifv2}.

\begin{table}[b]
\centering
\small
\setlength{\tabcolsep}{3.8pt}
\begin{tabular}{llr}
\toprule
Track & System & \favgtwo{} \\
\midrule
1 & Qwen 2.5 72B baseline & 73.66 \\
1 & Retrieved demos, Gemma-4-31B-IT & \textbf{85.22} \\
\midrule
2 & Qwen 2.5 72B baseline & 74.75 \\
2 & Local Gemma-4-31B-IT stack & 86.58 \\
2 & GPT-5.5 target-grounded classifier & 91.48 \\
2 & Full target-grounded pipeline & \textbf{91.78} \\
\bottomrule
\end{tabular}
\caption{Official test results on a 0 to 100 scale. Track 1 contains 352 Women
Driving tweets; Track 2 contains 644 tweets across two unseen targets. Baseline
values are reported by \citet{albalawi2026mawqifv2}.}
\label{tab:official}
\end{table}

Cross-target stance systems have represented targets explicitly
\citep{allaway2020tganet}, used contrastive objectives
\citep{liang2022jointcl}, and injected retrieved or generated knowledge
\citep{he2022wsbert,beck-etal-2023-robust,zhang-etal-2024-llm-driven}.
Stance Reasoner generates explicit chain-of-thought rationales and aggregates
self-consistent samples \citep{taranukhin-etal-2024-stance}. We instead separate a
reusable target description from constrained label prediction. Arabic
shared-task systems have used prompt engineering or LLM fine-tuning
\citep{hariri2024smash,alghaslan2024mgkm}. Unlike methods that retrieve
Wikipedia passages or generate tweet-specific knowledge, our target card is a
single neutral, target-only description reused for every tweet. We test this
alongside retrieved same-target examples as two lightweight forms of
inference-time target evidence.

Our public comparisons separately test exact-target labels, demonstration selection, factual context, and source-target adaptation. A committee analysis examines whether disagreement identifies errors, while a cross-corpus rubric study tests the transfer of annotation cues.

\section{Task and data}
\label{sec:data}

The original Mawqif splits contain 3,502 training and 619 development tweets
about COVID-19 Vaccine, Digital Transformation, and Women Empowerment. Each
tweet has stance, sentiment, and sarcasm labels. Track 1 evaluates 352 Women
Driving tweets. Track 2 evaluates 332 Electric Cars and 312 Trimester System
tweets. The primary metric is
\(\favgtwo=(F_{\textsc{Favor}}+F_{\textsc{Against}})/2\); \textsc{None} is not
averaged in its numerator \citep{mohammad2016semeval}.
We also report \favgthree{}, the macro-F1 over all three stance labels, and overall accuracy for the final systems.
Each instance pairs an Arabic tweet with a named target, and the system returns
exactly one of \mbox{\textsc{Favor}}, \mbox{\textsc{Against}}, or
\mbox{\textsc{None}}. For example,
a tweet celebrating women receiving driving licences should be classified as
\textsc{Favor} when paired with the Women Driving target.

For Track 1, we use ArabicStanceX \citep{arabicstancex} as an external
demonstration corpus; it is not part of the shared-task data. The corpus is
publicly released under CC BY 4.0. Its 947 Women Driving tweets have no
exact or Arabic-normalised text overlap with the 352 task tweets. The main
public ablation uses a fixed 497-row retrieval pool and a disjoint 450-row
evaluation set. A separate target-control study uses the official 756/191
train/test split. Official StanceEval inference retrieves from all 947 labelled
tweets. Separate public experiments evaluate target-held-out prompting and source-only adaptation on Mawqif.

\section{System overview}

\subsection{Track 1: retrieved in-target demonstrations}

For each task tweet, we retrieve the eight nearest Women Driving examples from
the full ArabicStanceX pool, following similarity-based demonstration selection
\citep{liu2022whatmakes}. Similarity is cosine similarity over character-boundary
TF-IDF features. We use character
features to reduce sensitivity to dialectal spelling variation and short social-media
text. The prompt contains the target, a factual background, the tweet, the
retrieved labelled examples, and a one-sentence instruction restricting the
answer to the three stance labels.

We run Gemma-4-31B-IT locally with vLLM
\citep{kwon2023vllm,gemmateam2026gemma4technicalreport}. Instead of generating
an explanation, the model scores the continuations \textsc{Favor},
\textsc{Against}, and \textsc{None}, and we select the largest log-probability.
The Track 1 system in Table~\ref{tab:official} performs no parameter update and
uses no class quota.

\subsection{Track 2: prompted factual target context}

\paragraph{Target descriptions.}
The pipeline caches two English descriptions for each unseen target. Gemini
3.6 Flash generates the card used by the GPT and Gemini classifiers, while
Gemma-4-31B-IT generates the card used by its own scorer. Both prompts request
two or three factual sentences about the topic in Saudi Arabia and the Arab
world, explicitly forbidding opinions. Each card is generated at temperature
zero and reused unchanged for every tweet about its target.
The stance prompt contains the Arabic and English target names, the card, and the tweet, and requests one stance label.

\paragraph{Classification and arbitration.}
GPT-5.5 supplies the initial target-grounded label, and Gemini 3.6 Flash labels
each row independently. GPT-5.5 Pro arbitrates their 44 disagreements, changing
13 initial labels. Gemma-4-31B-IT and a second GPT-5.5 pass independently score
the three label tokens.

\paragraph{Score fusion and assignment.}
For row \(i\) and class \(c\),
we combine these signals with the fixed log-linear rule
\begin{equation}
s_{i,c}=2\log p^{\mathrm{anchor}}_{i,c}
       +\log p^{\mathrm{Gemma}}_{i,c}
       +\log p^{\mathrm{GPT}}_{i,c}.
\label{eq:fusion}
\end{equation}
where the anchor assigns 0.85 to its selected class and 0.075 to each alternative. For each scorer, we softmax the three label log scores and floor probabilities at $10^{-12}$ before taking logs. A Hungarian assignment \citep{kuhn1955hungarian} maximises the total score subject to class totals derived from the official website.\footnote{``View Dataset Statistics'' at \url{https://stanceeval.github.io/\#data}; also present in the June 18, 2026 revision, before the test release.} Its anonymous test rows give sizes and \mbox{\textsc{Favor}}/\mbox{\textsc{Against}}/\mbox{\textsc{None}} percentages. Matching sizes to the released data identifies 332 tweets (43.67/35.24/21.08\%) as Electric Cars and 312 (22.12/72.44/5.45\%) as Trimester System. Multiplying by the respective sizes and rounding gives 145/117/70 and 69/226/17.

The Gemma scorer alone receives normalised Arabic text (NFKC, diacritic and tatweel removal, letter normalisation, and URL and mention placeholders); API classifiers receive the raw tweet. The anchor uses medium reasoning, arbitration uses high reasoning, and the score-only GPT pass disables reasoning. Fusion and count assignment complete the official pipeline; the controlled comparisons in Section~\ref{sec:results} use public held-out labels.

\section{Experimental setup}

All controlled comparisons use public labels and keep evaluation rows separate
from retrieval, training, and checkpoint selection. Track 1 uses the two
ArabicStanceX splits described in Section~\ref{sec:data}. Track 2 compares no-card,
fixed-card, and gated-card prompting under target-held-out Mawqif evaluation.

For each held-out target, the \emph{no-card} condition supplies only the target
and tweet. The \emph{fixed-card} condition prepends the same cached factual
description to every tweet about that target. The \emph{gated-card} condition
uses an independent Arabic yes/no prompt to decide whether to prepend that
description for each tweet. The Gemma-4-12B-IT comparison instead contrasts the
unchanged base checkpoint with a source-trained QLoRA adapter. The public card study uses Arabic Gemma-generated descriptions and one-word generation at temperature zero, not the official English-card scoring protocol.

\paragraph{Parameter adaptation.}
For parameter adaptation, we fine-tune Gemma-4-12B-IT with QLoRA
\citep{dettmers2023qlora}. Each fold trains on two Mawqif targets and evaluates
on the third. The rank-16
adapter uses scaling 32, dropout 0.05,
maximum length 256, effective batch size 16, and seed 42. A 15\% source-only
validation slice jointly selects the learning rate and checkpoint. The grid
contains \(5\times10^{-5}\) and \(10^{-4}\), with \(2\times10^{-4}\) also
tested for the COVID-19 Vaccine fold. The selected learning rates are
\(10^{-4}\) for Women Empowerment and Digital Transformation and
\(5\times10^{-5}\) for COVID-19 Vaccine; all selected checkpoints occur near
two-thirds of an epoch.

\section{Results}
\label{sec:results}

Table~\ref{tab:official} reports the official results. The final systems also score 67.92 \favgthree{} and 81.25 accuracy on Track 1, and 83.53 and 88.82 on Track 2, pooled over both targets. The Track 1 retrieval system scores 85.22 \favgtwo{}, ranking sixth among 21 teams; the complete Track 2 pipeline scores 91.78, ranking fourth among 13 teams in the final standings. Its target scores are 88.75 on Electric Cars and 95.23 on Trimester System. A local-only Track 2 comparison combines eight Gemma-4-31B-IT prompt views with a categorical stacker fitted on public Women Empowerment labels and scores 86.58. The direct GPT-5.5 classifier scores 91.48. This final-test comparison is descriptive; the 0.30-point difference is not a component ablation.

\begin{table}[t]
\centering
\footnotesize
\setlength{\tabcolsep}{1.6pt}
\begin{tabular}{lrr}
\toprule
\multicolumn{3}{l}{\emph{Track 1, held-out ArabicStanceX}} \\
Variant & Public & Official test \\
\midrule
direct, no demonstrations & 79.45 & 79.51 \\
8 random target demos & 84.78 & 83.85 \\
8 retrieved target demos & \textbf{89.04} & \textbf{85.22} \\
\bottomrule
\end{tabular}

\vspace{3pt}
\begin{tabular}{llrrrr}
\toprule
\multicolumn{6}{l}{\emph{Track 2, target-held-out Mawqif}} \\
Model & Variant & WE & CV & DT & Mean \\
\midrule
Gemma-4-31B-IT & no card & 87.95 & 92.31 & \textbf{79.37} & 86.54 \\
Gemma-4-31B-IT & fixed card & \textbf{89.25} & \textbf{92.34} & 79.20 & \textbf{86.93} \\
Gemma-4-31B-IT & gated card & 88.51 & 92.25 & 79.20 & 86.65 \\
Gemma-4-12B-IT & base & \textbf{86.18} & \textbf{88.51} & \textbf{78.60} & \textbf{84.43} \\
Gemma-4-12B-IT & QLoRA & 86.10 & 87.50 & 77.70 & 83.77 \\
\bottomrule
\end{tabular}
\caption{\favgtwo{} on a 0 to 100 scale. The Track 1 public column evaluates
450 ArabicStanceX tweets with a disjoint 497-row demonstration pool; the
official-test column reports the same three submitted arms descriptively and is
not used for component selection. WE, CV, and DT denote Women Empowerment,
COVID-19 Vaccine, and Digital Transformation. Track 2 columns are separate
public held-out targets. Bold marks the best result within each model block.}
\label{tab:heldout}
\end{table}

\paragraph{Target demonstrations and their selection both help.}
On the 450 public rows, random demonstrations improve on direct prompting, and retrieval improves further (Table~\ref{tab:heldout}). Exact McNemar tests on paired correctness reject equal error rates for
both transitions ($p=1.4\times10^{-5}$ and $p=1.6\times10^{-3}$). The
\textsc{None} F1 rises from 25.29 to 52.00. The official-test arms preserve the ordering but are descriptive, not an ablation. A separate 191-row public control uses fixed nine-example sets with three examples per stance, holding the corpus and annotation convention fixed: direct prompting scores 71.57; nine Women Driving
examples score 81.84, compared with 73.10 for nine other-target ArabicStanceX
examples (difference 8.74; 95\% paired-bootstrap CI [4.40, 13.38]). Together,
the comparisons distinguish
access to target labels from selecting locally relevant examples.

\paragraph{Target cards and source-only adaptation.}
The fixed card's change is target-dependent: $+1.30$ on Women
Empowerment, $+0.03$ on COVID-19 Vaccine, and $-0.17$ on Digital
Transformation. The gated-card condition selects the card for 327 of 4,121
rows (7.9\%) and reaches 86.65, below the fixed-card condition. Neither is a matched ablation of the submitted GPT-5.5 anchor. QLoRA loses on every 12B fold. This finding is specific to the tested recipe, and the
31B and 12B blocks are within-model comparisons rather than a shared-backbone
comparison.

\section{Error analysis}

On public Women Empowerment data, a committee of GPT-5.6 Luna, GPT-5.6 Terra,
and Gemma-4-31B-IT is wrong on 54 of 136 split votes (39.7\%), but only 99 of
1,264 unanimous votes (7.8\%). A five-fold logistic stack reaches 90.08
\favgtwo{}, below the 90.38 of plain majority. Disagreement is therefore a risk
marker, not a correction rule.

Qualitative inspection points to \emph{stance attachment}. One tweet jokes
that, ``in the era of women's empowerment, they will not do anything to you'';
the three models produce all three labels, while the annotation is
\textsc{None}. Local affect is easier to recognise than whether it is directed
at the named target.

\paragraph{A corpus-specific rubric does not transfer.}
Post hoc, we distilled a 60-line Arabic rubric from recurring Women Driving
evaluation-corpus patterns, including sarcasm, hashtag reversal, celebration,
supplication, and autonomy cues. Because its construction used inspection of
that corpus, it was not part of the reported system; we test only its transfer
to held-out ArabicStanceX Women Driving labels. Even after removing every
verbatim fragment, the detailed rubric underperforms a four-line minimal
instruction for a two-model ensemble (87.67 versus 89.18 \favgtwo{}) and for
both members (87.02 versus 90.11; 87.42 versus 88.04). The tested corpus-derived rubric did not improve cross-corpus prediction over the minimal instruction.

\begin{table}[t]
\centering
\small
\setlength{\tabcolsep}{5pt}
\begin{tabular}{lrr}
\toprule
Target & Track & Fleiss' $\kappa$ \\
\midrule
Women Driving & 1 & 0.479 \\
Electric Cars & 2 & 0.479 \\
Trimester System & 2 & 0.711 \\
\midrule
Overall & Both & 0.564 \\
\bottomrule
\end{tabular}
\caption{Human stance-label agreement in Mawqif-XT, as reported by
\citet{albalawi2026mawqifv2}.}
\label{tab:agreement}
\end{table}

The rubric result is consistent with a difficult \textsc{None} boundary rather
than missing target vocabulary alone. Table~\ref{tab:agreement} shows lower
human agreement for Women Driving and Electric Cars than for the Trimester
System. Across the three targets, 66.12\% of stance disagreements contrast
\textsc{None} with a polar label. These statistics do not establish a cause,
but they place the transfer failure in context: lexical cues alone may not
encode a corpus's annotation policy.

\section{Conclusion}

Our systems supply target information at inference time and reach 85.22 and 91.78 \favgtwo{}. Separate public comparisons support same-target demonstrations and retrieval, show a small target-dependent card effect, and find no benefit from the tested adapter. Target-shift methods should be selected on target-disjoint data. Model disagreement localises difficult rows but does not identify their correct labels.

\section*{Limitations}

Track 1 uses external labelled examples for its evaluation target and is not
directly comparable with task-data-only systems. We fix $k=8$ and do not
compare dense retrieval or demonstration order; ArabicStanceX also differs
from Mawqif in annotation and class balance. The target-card study covers three
targets, and its 0.39-point mean gain lacks a confidence interval. Track 2 lacks
target-held-out component ablations, while count assignment assumes known
label totals. The post-hoc rubric study covers one target and tests cross-corpus
transfer. Hosted-model drift limits repeatability.

\section*{Acknowledgments}

We thank the StanceEval organizers, dataset annotators, and anonymous reviewer for their contributions and feedback.

\clearpage
\bibliography{refs}

@inproceedings{alturayeif2022mawqif,
  title     = {{Mawqif}: A Multi-label {A}rabic Dataset for Target-specific Stance Detection},
  author    = {Alturayeif, Nora Saleh and Luqman, Hamzah Abdullah and Ahmed, Moataz Aly Kamaleldin},
  booktitle = {Proceedings of the Seventh Arabic Natural Language Processing Workshop (WANLP)},
  year      = {2022},
  address   = {Abu Dhabi, United Arab Emirates},
  publisher = {Association for Computational Linguistics},
  doi       = {10.18653/v1/2022.wanlp-1.16},
  pages     = {174--184}
}

@inproceedings{mohammad2016semeval,
  title = {{SemEval}-2016 Task 6: Detecting Stance in Tweets},
  author = {Mohammad, Saif and Kiritchenko, Svetlana and Sobhani, Parinaz and Zhu, Xiaodan and Cherry, Colin},
  booktitle = {Proceedings of the 10th International Workshop on Semantic Evaluation (SemEval-2016)},
  year = {2016},
  address = {San Diego, California},
  publisher = {Association for Computational Linguistics},
  url = {https://aclanthology.org/S16-1003/},
  doi = {10.18653/v1/S16-1003},
  pages = {31--41}
}

@inproceedings{allaway2020tganet,
  title = {Zero-Shot Stance Detection: A Dataset and Model using Generalized Topic Representations},
  author = {Allaway, Emily and McKeown, Kathleen},
  booktitle = {Proceedings of the 2020 Conference on Empirical Methods in Natural Language Processing (EMNLP)},
  year = {2020},
  address = {Online},
  publisher = {Association for Computational Linguistics},
  url = {https://aclanthology.org/2020.emnlp-main.717/},
  doi = {10.18653/v1/2020.emnlp-main.717},
  pages = {8913--8931}
}

@inproceedings{liang2022jointcl,
  title     = {{JointCL}: A Joint Contrastive Learning Framework for Zero-Shot Stance Detection},
  author    = {Liang, Bin and Zhu, Qinglin and Li, Xiang and Yang, Min and Gui, Lin and He, Yulan and Xu, Ruifeng},
  booktitle = {Proceedings of the 60th Annual Meeting of the Association for Computational Linguistics (Volume 1: Long Papers)},
  year      = {2022},
  address   = {Dublin, Ireland},
  publisher = {Association for Computational Linguistics},
  doi       = {10.18653/v1/2022.acl-long.7},
  pages     = {81--91}
}

@inproceedings{taranukhin-etal-2024-stance,
  title     = {Stance Reasoner: Zero-Shot Stance Detection on Social Media with Explicit Reasoning},
  author    = {Taranukhin, Maksym and Shwartz, Vered and Milios, Evangelos},
  booktitle = {Proceedings of the 2024 Joint International Conference on Computational Linguistics, Language Resources and Evaluation (LREC-COLING 2024)},
  month     = may,
  year      = {2024},
  address   = {Torino, Italia},
  publisher = {ELRA and ICCL},
  url       = {https://aclanthology.org/2024.lrec-main.1326/},
  pages     = {15257--15272}
}

@inproceedings{he2022wsbert,
  title     = {Infusing Knowledge from {W}ikipedia to Enhance Stance Detection},
  author    = {He, Zihao and Mokhberian, Negar and Lerman, Kristina},
  booktitle = {Proceedings of the 12th Workshop on Computational Approaches to Subjectivity, Sentiment {\&} Social Media Analysis (WASSA)},
  year      = {2022},
  address   = {Dublin, Ireland},
  publisher = {Association for Computational Linguistics},
  doi       = {10.18653/v1/2022.wassa-1.7},
  pages     = {71--77}
}

@inproceedings{liu2022whatmakes,
  title     = {What Makes Good In-Context Examples for {GPT}-3?},
  author    = {Liu, Jiachang and Shen, Dinghan and Zhang, Yizhe and Dolan, Bill and Carin, Lawrence and Chen, Weizhu},
  booktitle = {Proceedings of Deep Learning Inside Out (DeeLIO 2022): The 3rd Workshop on Knowledge Extraction and Integration for Deep Learning Architectures},
  year      = {2022},
  address   = {Dublin, Ireland and Online},
  publisher = {Association for Computational Linguistics},
  doi       = {10.18653/v1/2022.deelio-1.10},
  pages     = {100--114}
}

@article{arabicstancex,
  title   = {Constructing and evaluating {ArabicStanceX}: a social media dataset for {A}rabic stance detection},
  author  = {Alkhathlan, Ali and Alahmadi, Faris and Kateb, Faris and Al-Khalifa, Hend},
  journal = {Frontiers in Artificial Intelligence},
  volume  = {8},
  pages   = {1615800},
  year    = {2025},
  doi     = {10.3389/frai.2025.1615800}
}

@misc{albalawi2026mawqifv2,
      title={{Mawqif-XT}: An {A}rabic Benchmark Dataset for Cross-Target Stance Detection},
      author={Rasha Albalawi and Nuha Albadi and Hamzah Luqman and Maram Kurdi and Saad Ezzini and Asma Yamani and Ahmed Ashraf},
      year={2026},
      eprint={2608.09539},
      archivePrefix={arXiv},
      primaryClass={cs.CL},
      doi={10.48550/arXiv.2608.09539},
      url={https://arxiv.org/abs/2608.09539}
}

@inproceedings{albalawi2026stanceeval,
    title = {{StanceEval}-2026: The Second Stance Detection Shared Task},
    author = {Albalawi, Rasha and Albadi, Nuha and Luqman, Hamzah and Ezzini, Saad and Kurdi, Maram and Yamani, Asma and Ashraf, Ahmed and Al-Shaibani, Maged and Alturayeif, Nora},
    booktitle = {Proceedings of the Fourth Arabic Natural Language Processing Conference (ArabicNLP 2026)},
    year = {2026},
    address = {Budapest, Hungary},
    publisher = {Association for Computational Linguistics}
}

@inproceedings{hariri2024smash,
  title     = {{SMASH} at {StanceEval} 2024: Prompt Engineering {LLM}s for {A}rabic Stance Detection},
  author    = {Al Hariri, Youssef and Abu Farha, Ibrahim},
  booktitle = {Proceedings of the Second Arabic Natural Language Processing Conference},
  year      = {2024},
  address   = {Bangkok, Thailand},
  publisher = {Association for Computational Linguistics},
  url       = {https://aclanthology.org/2024.arabicnlp-1.92/},
  doi       = {10.18653/v1/2024.arabicnlp-1.92},
  pages     = {800--806}
}

@article{kuhn1955hungarian,
  title   = {The {H}ungarian method for the assignment problem},
  author  = {Kuhn, Harold W.},
  journal = {Naval Research Logistics Quarterly},
  volume  = {2},
  number  = {1-2},
  pages   = {83--97},
  year    = {1955},
  doi     = {10.1002/nav.3800020109}
}

@inproceedings{dettmers2023qlora,
  title     = {{QLoRA}: Efficient Finetuning of Quantized {LLM}s},
  author    = {Dettmers, Tim and Pagnoni, Artidoro and Holtzman, Ari and Zettlemoyer, Luke},
  booktitle = {Advances in Neural Information Processing Systems (NeurIPS)},
  volume    = {36},
  pages     = {10088--10115},
  publisher = {Curran Associates, Inc.},
  year      = {2023},
  url       = {https://proceedings.neurips.cc/paper_files/paper/2023/hash/1feb87871436031bdc0f2beaa62a049b-Abstract-Conference.html},
  doi       = {10.52202/075280-0441}
}

@inproceedings{alghaslan2024mgkm,
  title     = {{MGKM} at {StanceEval}2024 Fine-Tuning Large Language Models for {A}rabic Stance Detection},
  author    = {Alghaslan, Mamoun and Almutairy, Khaled},
  booktitle = {Proceedings of the Second Arabic Natural Language Processing Conference},
  year      = {2024},
  address   = {Bangkok, Thailand},
  publisher = {Association for Computational Linguistics},
  url       = {https://aclanthology.org/2024.arabicnlp-1.95/},
  doi       = {10.18653/v1/2024.arabicnlp-1.95},
  pages     = {816--822}
}

@inproceedings{kwon2023vllm,
  title     = {Efficient Memory Management for Large Language Model Serving with {PagedAttention}},
  author    = {Kwon, Woosuk and Li, Zhuohan and Zhuang, Siyuan and Sheng, Ying and Zheng, Lianmin and Yu, Cody Hao and Gonzalez, Joseph E. and Zhang, Hao and Stoica, Ion},
  booktitle = {Proceedings of the 29th Symposium on Operating Systems Principles (SOSP)},
  year      = {2023},
  pages     = {611--626},
  doi       = {10.1145/3600006.3613165}
}

@inproceedings{zhang-etal-2024-llm-driven,
  title     = {{LLM}-Driven Knowledge Injection Advances Zero-Shot and Cross-Target Stance Detection},
  author    = {Zhang, Zhao and Li, Yiming and Zhang, Jin and Xu, Hui},
  booktitle = {Proceedings of the 2024 Conference of the North American Chapter of the Association for Computational Linguistics: Human Language Technologies (Volume 2: Short Papers)},
  year      = {2024},
  address   = {Mexico City, Mexico},
  publisher = {Association for Computational Linguistics},
  url       = {https://aclanthology.org/2024.naacl-short.32/},
  doi       = {10.18653/v1/2024.naacl-short.32},
  pages     = {371--378}
}

@inproceedings{beck-etal-2023-robust,
  title     = {Robust Integration of Contextual Information for Cross-Target Stance Detection},
  author    = {Beck, Tilman and Waldis, Andreas and Gurevych, Iryna},
  booktitle = {Proceedings of the 12th Joint Conference on Lexical and Computational Semantics (*SEM 2023)},
  year      = {2023},
  address   = {Toronto, Canada},
  publisher = {Association for Computational Linguistics},
  url       = {https://aclanthology.org/2023.starsem-1.43/},
  doi       = {10.18653/v1/2023.starsem-1.43},
  pages     = {494--511}
}

@misc{gemmateam2026gemma4technicalreport,
  title = {Gemma 4 Technical Report},
  author = {{Gemma Team}},
  year = {2026},
  eprint = {2607.02770},
  archivePrefix = {arXiv},
  primaryClass = {cs.CL},
  doi = {10.48550/arXiv.2607.02770},
  url = {https://arxiv.org/abs/2607.02770}
}

\end{document}